\documentclass[11pt]{article}

\usepackage{acl}

\usepackage{times}
\usepackage{latexsym}

\usepackage[T1]{fontenc}

\usepackage[utf8]{inputenc}

\usepackage{microtype}

\usepackage{inconsolata}

\usepackage{graphicx}
\usepackage{amsmath}
\usepackage{amsfonts}
\usepackage{amssymb}
\usepackage{booktabs}
\usepackage{multirow}
\usepackage{hyperref}
\usepackage{tcolorbox}

\title{Interpretable Graph-Language Modeling for Detecting \\Youth Illicit Drug Use}

\author{
 \textbf{Yiyang Li\textsuperscript{1*}},
 \textbf{Zehong Wang\textsuperscript{1*}},
 \textbf{Zhengqing Yuan\textsuperscript{1}},
 \textbf{Zheyuan Zhang\textsuperscript{1}},
\\
 \textbf{Keerthiram Murugesan\textsuperscript{2}},
 \textbf{Chuxu Zhang\textsuperscript{3}},
 \textbf{Yanfang Ye\textsuperscript{1\textdagger}}
\\
 \textsuperscript{1}University of Notre Dame,
 \textsuperscript{2}IBM Research,
 \textsuperscript{3}University of Connecticut\\
\textsuperscript{*}Equal Contribution \quad 
\textsuperscript{\textdagger}Corresponding Author\\
\texttt{\{yli62, yye7\}@nd.edu}
}

\begin{document}

\maketitle

\begin{abstract}
    
    Illicit drug use among teenagers and young adults (TYAs) remains a pressing public health concern, with rising prevalence and long-term impacts on health and well-being. To detect illicit drug use among TYAs, researchers analyze large-scale surveys such as the Youth Risk Behavior Survey (YRBS) and the National Survey on Drug Use and Health (NSDUH), which preserve rich demographic, psychological, and environmental factors related to substance use. However, existing modeling methods treat survey variables independently, overlooking latent and interconnected structures among them. To address this limitation, we propose \textbf{LAMI} (\textit{LAtent relation Mining with bi-modal Interpretability}), a novel joint graph-language modeling framework for detecting illicit drug use and interpreting behavioral risk factors among TYAs. LAMI represents individual responses as relational graphs, learns latent connections through a specialized graph structure learning layer, and integrates a large language model to generate natural language explanations grounded in both graph structures and survey semantics. Experiments on the YRBS and NSDUH datasets show that LAMI outperforms competitive baselines in predictive accuracy. Interpretability analyses further demonstrate that LAMI reveals meaningful behavioral substructures and psychosocial pathways, such as family dynamics, peer influence, and school-related distress, that align with established risk factors for substance use. Our codebase is available \href{https://anonymous.4open.science/r/LAMI-5AEC/README.md}{here}.

\end{abstract}

\section{Introduction}

Illicit drug use~\cite{cdc_illicitDrugUse} represents a persistent and escalating public health challenge in the United States, contributing to significant social and health burdens. It includes both the consumption of illegal substances, such as cannabis, cocaine, and hallucinogens, and the misuse of prescription medications. Teenagers and young adults (TYAs) are particularly at risk~\cite{simon2022adolescent}, exhibiting disproportionately high prevalence rates and associated harms. According to the 2024 \textit{National Survey on Drug Use and Health} (NSDUH), 38.1\% of young adults aged 18–25 and 15.1\% of adolescents aged 12–17 reported illicit drug use~\cite{samhsa_nsduh2025}. These alarming statistics highlight an urgent need to focus on TYAs to better understand the behavioral and psychosocial risk factors that contribute to illicit drug use, thereby informing more effective detection and prevention strategies.

Understanding why TYAs are particularly vulnerable is critical for developing effective detection and prevention strategies. Biologically, during adolescence and emerging adulthood, the brain’s reward system matures earlier than the executive control system, heightening reward sensitivity while limiting impulse regulation~\cite{steinberg2008social, luciana2013adolescent}. This developmental imbalance increases susceptibility to risk-taking behaviors, including illicit drug use~\cite{geier2013adolescent, romer2017beyond}. At the same time, TYAs experience major life transitions - such as leaving home, pursuing higher education, or entering the workforce -while coping mechanisms and identity formation are still developing~\cite{arnett2000emerging, patton2016our}. Together, these interacting biological, psychological, and social factors contribute to elevated risk and underscore the need to identify the key mechanisms driving substance use vulnerability among TYAs.

\begin{figure*}[t]
\centering
\includegraphics[width=\textwidth]{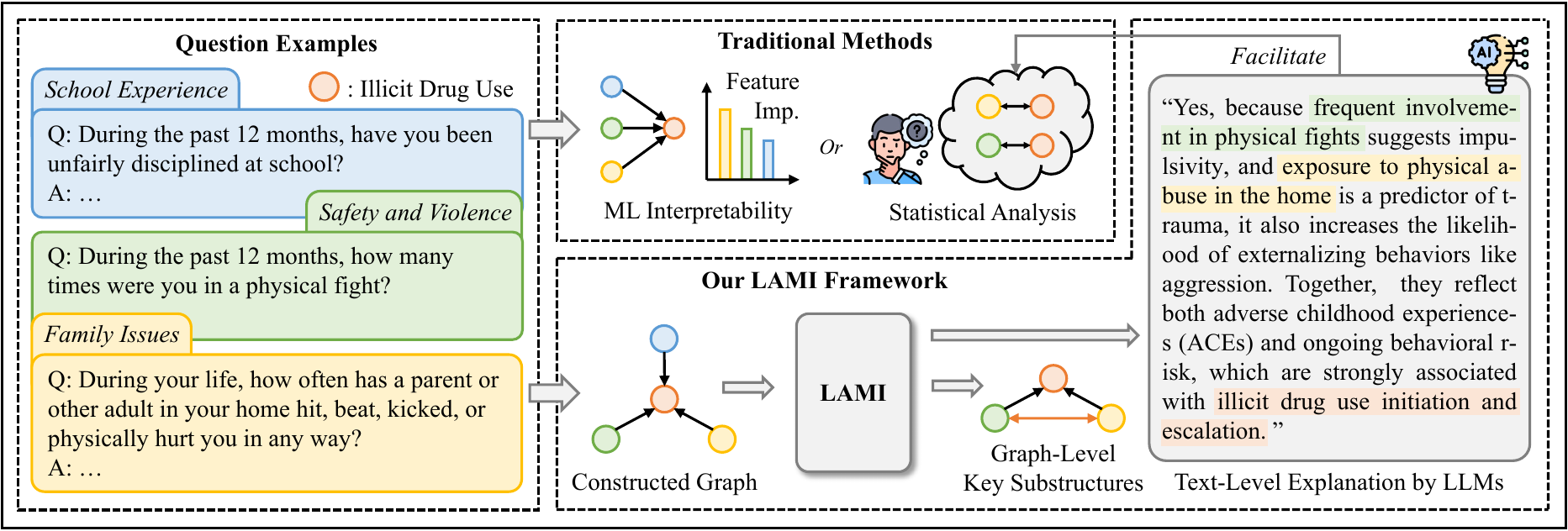}
\vspace{-10pt}
\caption{\textbf{Comparison between traditional methods and our LAMI for illicit drug use detection.} Traditional approaches typically treat survey variables in isolation, relying on feature importance metrics or expert-selected variables, which overlook the interdependent relationships among behavioral and psychosocial factors. In contrast, LAMI models survey responses as relational graphs, enabling the discovery of latent structural risk factors and facilitating deeper, more interpretable behavioral insights through graph-level analysis and LLM-based explanations.}
\label{fig1}
\vspace{-10pt}
\end{figure*}

To leverage these multifaceted risk factors for detecting illegal drug use among TYAs, researchers have long relied on large-scale surveys, such as the \textit{Youth Risk Behavior Survey} (YRBS)~\cite{CDC_YRBSS} and the \textit{National Survey on Drug Use and Health} (NSDUH)~\cite{NSDUH}, that capture diverse demographic, psychosocial, behavioral, and environmental variables. Traditional approaches ranging from statistical models to conventional machine learning techniques~\cite{wadekar2019predicting, han2020using, thompson2021clinical, mckinnon2024experiences, kim2025machine} have offered valuable insights, but they typically treat survey variables independently. This assumption overlooks the complex and interdependent relationships that shape behavioral outcomes, thereby obscuring latent structures that may reveal deeper mechanisms of risk. As illustrated in Figure~\ref{fig1}, uncovering these latent relations can enable a more holistic understanding of how different factors interact and provide new pathways for explaining and detecting TYAs illicit drug use.

Building on this motivation, our work aims to model the relational information embedded within survey data to uncover the underlying behavioral and psychosocial structures associated with illicit drug use. However, capturing these relations presents several key challenges. First, survey data are not inherently relational, making it difficult to represent variable-level dependencies in a suitable form. Second, the latent relations among variables are not explicitly observed and must be inferred through specialized learning methods. Third, predictive frameworks must remain interpretable: able to identify meaningful substructures and generate explanations that support researchers and practitioners in public health and behavioral science.


    


To address these challenges, we propose \textbf{LA}tent relation \textbf{M}ining with bi-modal \textbf{I}nterpretability (\textbf{LAMI}), a novel framework that integrates graph neural networks (GNNs)~\cite{khemani2024review} with large language models (LLMs)~\cite{zhao2023survey} for illicit drug use detection among TYAs. Specifically, LAMI (1) innovatively represents each individual’s survey responses as a relational graph, (2) learns latent relations among question nodes through a relational graph structure learning layer and a self-supervised pretext task, and (3) enhances interpretability by coupling GNN-derived substructure analysis with LLM-generated natural language explanations. We evaluate LAMI on the YRBS and NSDUH, demonstrating its superior predictive performance and interpretability. 

\section{Related Works}

\paragraph{Illicit Drug Use Prediction.}
Traditional statistical and machine learning approaches have long dominated the prediction of illicit drug use~\cite{unlu2025substance}. Valued for their simplicity and interpretability~\cite{king2017common}, these methods have identified key behavioral and psychosocial predictors, including smoking, alcohol use, mental health indicators (e.g., depression and anxiety), and suicidal ideation~\cite{wadekar2019predicting, han2020using, islam2023machine, rakovski2023predictors, kim2025machine}. Such findings have contributed substantially to public health research by informing prevention and intervention strategies. 
More recently, deep learning models have been applied to unstructured data sources such as electronic health records (EHR)~\cite{dong2021predicting, kashyap2023deep}, dietary patterns~\cite{zhang2024diet}, and social media~\cite{dou2021harnessing, ma2025llm}, demonstrating improved predictive accuracy and potential for digital surveillance. However, their use in large-scale population surveys remains limited, leaving research gap for modeling the complex relational dependencies that underlie behavioral and psychosocial factors in such data.

\paragraph{Graph Neural Networks and Structure Learning.}
GNNs~\citep{kipf2017semisupervised,hamilton2017inductive,velivckovic2017graph} have emerged as powerful tools for modeling structured data in which relationships among entities play a central role~\cite{wu2020comprehensive}. Recent work has extended GNNs to tabular settings, where explicit graph structures are not readily available~\cite{goodge2022lunar, du2022learning, telyatnikov2023egg, yan2023t2g}. 
However, their effectiveness critically depends on the quality of the manually defined graph structure - poorly constructed graphs can substantially degrade performance~\cite{zhu2021survey}. 
To mitigate this issue, Graph Structure Learning (GSL) approaches have been developed to learn or refine graph topologies during training~\cite{franceschi2019learning, jin2020graph, zhao2021heterogeneous, liu2022towards}. Nonetheless, most GSL methods do not explicitly account for the context of node attributes when modeling structural relationships, thereby limiting the ability to capture the nuanced inter-variable dependencies inherent in behavioral and psychosocial survey data.

\paragraph{Enhancing GNN Interpretability with LLMs.}
Recent research has explored integrating LLMs with GNNs to enhance interpretability, which can be broadly categorized into three paradigms. The first leverages LLMs to augment existing GNN explainers, where the LLM verbalizes, refines, or contextualizes explanations generated by traditional explainer modules~\cite{cedro2024graphxain, zhang2024llmexplainer, pan2024graphnarrator}. The second employs LLMs purely as external interpreters, providing post-hoc textual explanations without participating in model training or gradient propagation~\cite{zhang2024diet, he2024explaining}. The third paradigm achieves tighter coupling by injecting graph embeddings into LLM prompts as prefix tokens, aligning GNN and LLM representations and enabling the LLM to condition its reasoning on graph-encoded structural information~\cite{baghershahi2025nodes}. This deeper integration supports joint structural–semantic reasoning, which is particularly well-suited to our setting, where survey variables carry rich semantic meaning and their latent relations are essential for uncovering key psychosocial and behavioral substructures.

\section{Methodology}

\paragraph{Notations.}
Let \( G = (V, E, \mathcal{T}, \mathcal{R}) \) denote a relational graph, where \( V \) is the set of nodes, \( E \subseteq V \times V \) is the set of edges, \( \mathcal{T} \) is the set of node types, and \( \mathcal{R} \) is the set of edge types. Let \( X \in \mathbb{R}^{|V| \times d} \) be the node feature matrix. 

\paragraph{Problem Definition.}
Given a relational graph \( G \) constructed for a single TYA respondent, the objective is to learn a predictive function \( f \) that determines whether the individual engages in illicit drug use. To enhance interpretability, \( f \) further leverages an LLM to generate natural language explanations grounded in the learned graph representations.

\subsection{Relational Graph Construction}

\begin{table}[h]
\centering
\small
\resizebox{\linewidth}{!}{
\begin{tabular}{lcc}
\toprule
\textbf{Attribute} & \textbf{\texttt{YRBS}} & \textbf{\texttt{NSDUH}} \\
\midrule
\# Graphs & 19,931 & 21,510 \\
\# Positives  & 6,985 & 7,005 \\
\# Negatives & 12,964 & 14,505 \\ \midrule
\# User Nodes Per Graph & 1 & 1 \\
\# Question Nodes Per Graph & 88 & 349 \\
\# Topic Nodes Per Graph & 17 & 10 \\
\# Unique Relations Per Graph & 466 & 1,079 \\
\bottomrule
\end{tabular}
}
\caption{Statistics of the constructed datasets. }
\label{tab:dataset_stats}
\end{table}

\noindent We utilize two publicly available datasets, the 2023 releases of the YRBS and the NSDUH, to construct individual-level relational graphs based on their official documentation. Each respondent is represented by a graph \( G \) comprising a user node, a set of question nodes linked to the user node, and a set of topic nodes connected to their corresponding question nodes, where \( \mathcal{T} = \{\text{User}, \text{Question}, \text{Topic}\} \). The edge type between a user node and a question node encodes the respondent’s answer to that question. The user node features capture demographic characteristics, while question and topic node features are initialized using BERT-based embeddings~\cite{devlin2019bert} of their textual descriptions. To align with our focus on TYAs, we restrict the sample to respondents aged 15–25. Summary statistics of the constructed datasets are provided in Table~\ref{tab:dataset_stats}, and additional details are included in Appendix~\ref{sec:data prep}.

\begin{figure*}[t]
\centering
\includegraphics[width=\textwidth]{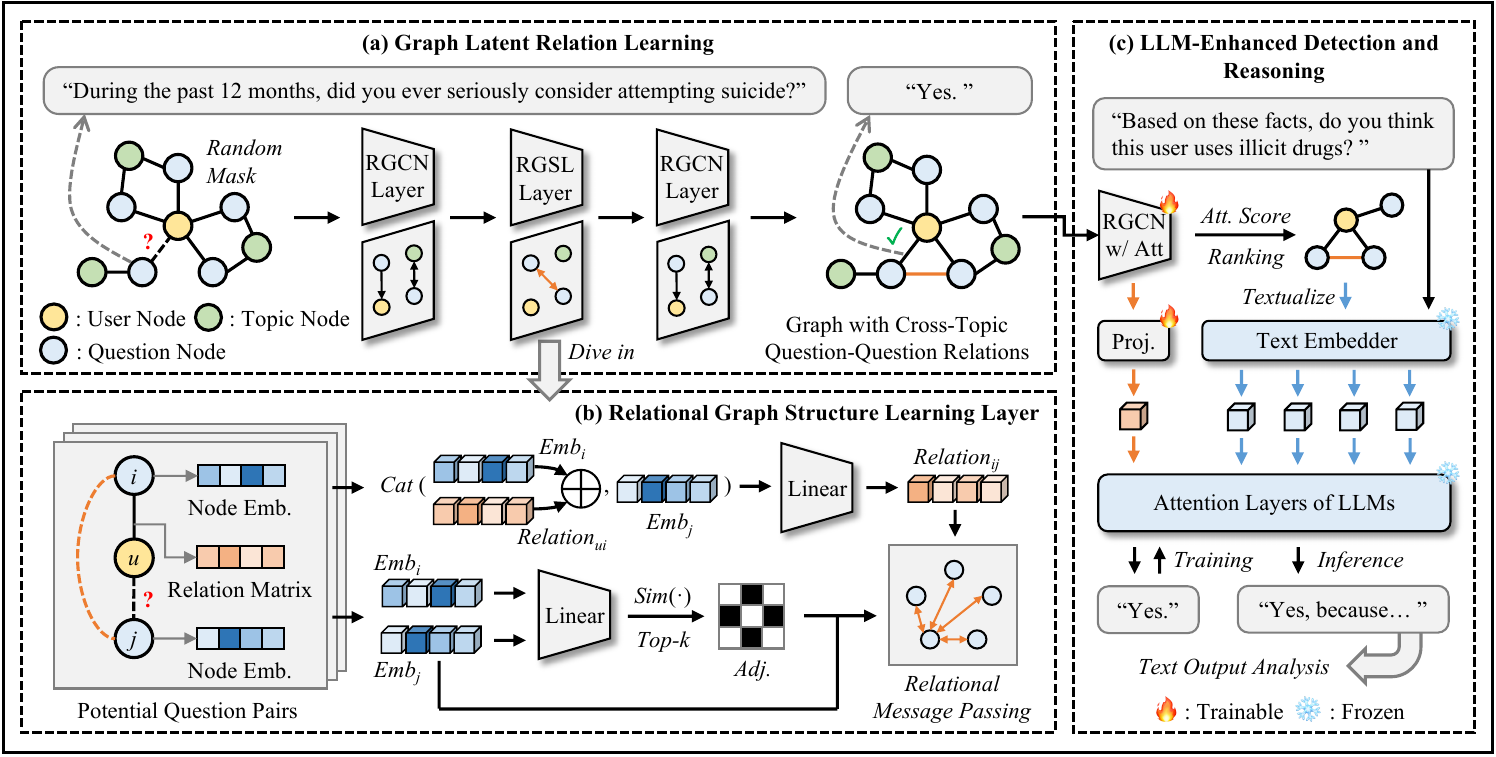}
\vspace{-10pt}
\caption{
\textbf{Overview of the LAMI framework.} (a) Graph latent relation learning for capturing implicit dependencies among survey variables, (b) a relational graph structure learning (RGSL) layer for inferring latent question–question connections, and (c) an LLM-enhanced predictor for detecting illicit drug use and generating text-based explanations. These components enable interpretable illicit drug use detection among TYAs.
}
\vspace{-10pt}

\label{fig2}
\end{figure*}

\subsection{Graph Latent Relation Learning}

Although many survey questions appear unrelated in isolation, they often reflect intertwined psychosocial factors that jointly lead to illicit drug use. For example, \emph{exposure to physical abuse in the home} may increase a respondent’s propensity for aggression and, consequently, a higher likelihood of \emph{frequent physical fights}. Taken together, these items jointly signal adverse childhood experiences (ACEs), which are strongly associated with drug use initiation. Uncovering such latent relations is therefore essential for both accurate prediction and interpretable analysis. Thus, the goal of this module is to refine the graph structure by discovering latent relations among nodes.

To model these hidden dependencies, we design a framework capable of discovering new relational structures while simultaneously performing representation learning. Specifically, our model stacks standard Relational Graph Convolutional Network (RGCN) layers with a \textit{Relational Graph Structure Learning (RGSL)} layer (Fig.~\ref{fig2}a). The RGCN layers follow the standard message-passing formulation~\cite{schlichtkrull2018modeling}:
\begin{equation}
h_i^{(l+1)} = \sigma \Big( \sum_{r \in \mathcal{R}} \sum_{j \in \mathcal{N}_i^r} \frac{1}{c_{i,r}} W_r^{(l)} h_j^{(l)} + W_0^{(l)} h_i^{(l)} \Big),
\end{equation}
where $\sigma(\cdot)$ is the activation function, $W_r^{(l)}$ is the learnable weight matrix for relation $r$ at layer $l$, $h_i^{(l)}$ denotes the embedding of node $i$, $\mathcal{N}_i^r$ represents the set of $r$-type neighbors, and $c_{i,r}$ is a normalization constant. The RGCN layers update node embeddings and learn user–question relations, while the RGSL layer leverages these learned relations as contextual signals to infer additional latent edges. 

\paragraph{RGSL Layer.}
The respondent’s answer to question $i$ often influences how question $i$ relates to another question $j$. Motivated by this observation, the RGSL layer is designed to induce context-dependent latent relations among question nodes. 
Given a target question node $i$ whose user–question edge $(u,j)$ is masked during training, we consider each other question node $j$ from a different topic. 
The layer (Fig.~\ref{fig2}b) (i) learns an adjacency matrix $A$ among question nodes and (ii) computes \emph{relation vectors} that inject user–question relation into cross-question interactions, before performing message passing over the learned structure. 

\emph{Adjacency matrix learning.} We first project question embeddings $h_i \in \mathbb{R}^d$ with a linear transformation $W_a \in \mathbb{R}^{d \times d}$ and compute pairwise scores $s_{ij}=(W_a h_i)^\top (W_a h_j)$ between any pairs of questions. Then select the top-$k_{\text{sim}}$ neighbors for each $i$ to form a sparse adjacency $A\in \{0,1\}^{Q\times Q}$ (with $Q$ questions). A hard mask is used in the forward pass, while gradients flow through $s_{ij}$. Additionally, we incorporate a penalty into the variance of the nodes’ degree to incentivize the model to discover more varied connections.

\emph{Relation vectors.} Let $r$ denote the known edge type between the user node $u$ and question node $i$, and let $W_r \in \mathbb{R}^{d\times d}$ be the corresponding relation matrix produced by the preceding RGCN layer. For each selected node $j$, we compute a relation vector
\begin{equation}
r_{ij} = \sigma \Big( W_r^{\mathrm{Rel}} \Big[ \big(h_i + \frac{1}{d}\!\sum_{k=1}^d (W_r)_{k}) \, \Big\| \, h_j \Big] \Big),
\end{equation}
where $W_r^{\mathrm{Rel}} \in \mathbb{R}^{2d \times d}$ is a trainable linear layer and $[\cdot\| \cdot]$ denotes concatenation. Injecting $W_r$ into $r_{ij}$ conditions the cross-question relation on the user–question relation context.

\emph{Message passing.} We update node $i$ by aggregating relation vectors from its learned neighbors:
\begin{equation}
h_i' = \sigma \Big( \sum_{j} \hat{A}_{ij}\, (r_{ij} \odot h_j) + W_s h_i \Big),
\end{equation}
where $\hat{A}$ is the row-normalized $A$, $\odot$ is element-wise multiplication, and $W_s \in \mathbb{R}^{d\times d}$ is a residual transformation. This equips the model to pass information only along newly discovered cross-topic edges that reflect latent relations. 

\paragraph{Self-Supervised Edge Type Prediction.}
To drive the discovery of latent relations, we adopt a self-supervised pretext task that predicts masked user–question edge types. For each graph, we randomly mask one edge $(u,q)$ (e.g., the user’s response to \emph{physical fight}) and train the model to infer its type from the remaining graph $G_{\setminus (u,q)}$, thereby encouraging it to capture dependencies such as the link between \emph{exposure to physical abuse} and \emph{physical fight}. Formally, letting $y_{(u,q)}^r\!\in\!\{0,1\}$ denote the true edge type and $P_r(u,q\mid G_{\setminus (u,q)})$ the predicted probability, we minimize
\begin{equation}
\mathcal{L}_{\text{edge}} = - \sum_{r \in \mathcal{R}} y_{(u,q)}^r\,\log P_r\!\left(u, q \mid G_{\setminus (u,q)}\right).
\end{equation}
During this pretext task, we restrict message flow through the user node: information from other questions cannot aggregate into $u$ to inform the prediction. Consequently, the model must rely on the RGSL-induced question–question edges to route information to the masked target, ensuring that improvements only stem from \emph{newly learned} latent structure. After this stage, we update the graph structure according to the learned adjacency matrix \( A \), forming an enriched relational graph.

\subsection{LLM-Enhanced Detection and Reasoning}

To identify key substructures and gain human-readable insights related to the detection results, we develop a bi-modal module that combines graph-based and language-based reasoning. As illustrated in Figure~\ref{fig2}~(c), the integration of GNN and LLM enables both structure-aware prediction and natural language explanation.

We first apply a standard RGCN model to compute node embeddings over the updated graph induced from the previous section: \( \mathbf{H} = \text{RGCN}(G) \in \mathbb{R}^{|V| \times d}\). To aggregate the information relevant to the user node, we introduce an attention mechanism on the question nodes, where the attention scores reflect the contribution of each node $\alpha_{i,u}=\operatorname{softmax}_{i\in\mathcal{N}_u}\!\left(\mathrm{MLP}\!\big([\mathbf{h}_i \,\|\, \mathbf{h}_u]\big)\right)$, 
with $\mathbf{h}_i,\mathbf{h}_u\in\mathbb{R}^{d}$ and a learnable scorer $\mathrm{MLP}$. The user embedding is updated via attention-weighted aggregation:
\begin{equation}
\mathbf{h}_u^{\text{agg}} = \sum_{i \in \mathcal{N}_u} \alpha_{i,u} \cdot \mathbf{h}_i,
\end{equation}

\paragraph{Textualization of Key Substructures.}
Based on $\alpha_{i,u}$, we select the top-$k_{\text{att}}$ question nodes and convert them to text using the original question wording and the user’s response. We also provide cues about learned interdependencies among selected questions. A short prompt then asks the LLM to predict whether the user uses illicit drugs. The resulting tokens are embedded by the LLM to obtain text embeddings $\mathcal{Z}=[\mathbf{z}_{t_1},\ldots,\mathbf{z}_{t_k}]$.

\begin{table*}[ht]
\centering
\resizebox{\linewidth}{!}{
\begin{tabular}{l ccccc ccccc}
\toprule
\multirow{2}{*}{Method} & \multicolumn{5}{c}{\texttt{YRBS}} & \multicolumn{5}{c}{\texttt{NSDUH}} \\
\cmidrule(lr){2-6} \cmidrule(lr){7-11}
 & Accuracy & Precision & Recall & F1-Score & AUC & Accuracy & Precision & Recall & F1-Score & AUC \\
\midrule
LightGBM    & 80.03 & 78.17 & 78.79 & 78.45 & \underline{86.70} & 79.95 & \underline{77.57} & \textbf{80.62} & \underline{78.35} & \underline{87.91} \\
XGBoost     & 79.93 & 78.21 & 77.50 & 77.82 & 85.59 & \underline{80.20} & 77.46 & \underline{79.75} & 78.24 & 87.52\\
\midrule
MLP         & 75.05 & 72.80 & 71.56 & 72.05 & 76.22 & 76.73 & 74.13 & 69.29 & 70.61 & 79.61 \\
TabNet      & 80.07 & 78.22 & 78.43 & 78.32 & 78.43 & 75.36 & 72.51 & 74.65 & 73.12 & 74.65 \\
\midrule
RGCN        & 80.29 & \underline{78.56} & 78.74 & \underline{78.65} & 86.57 & 79.29 & 76.44 & 78.90 & 77.24 & 87.14 \\
RGAT        & 79.42 & 77.74 & \underline{79.07} & 78.21 & 86.53 & 78.36 & 75.94 & 79.07 & 76.64 & 87.32 \\
\midrule
Qwen3-8B    & & & & & & & & & & \\
- \textit{Zero-shot} & 65.08 & 70.04 & 50.12 & 39.70 & 50.12 & 68.23 & 34.11 & 53.42 & 40.61 & 50.03 \\
- \textit{LoRA}      & \underline{82.20} & 78.52 & 67.71 & 72.68 & 78.93 & 78.51 & 69.74 & 57.29 & 62.90 & 72.80 \\
\midrule
\textbf{LAMI (Ours)} & \textbf{82.21} & \textbf{80.39} & \textbf{80.78} & \textbf{80.57} & \textbf{88.45} & \textbf{81.47} & \textbf{78.58} & 79.29 & \textbf{78.86} & \textbf{88.23} \\
\bottomrule
\end{tabular}
}
\caption{Performance comparison of different methods on \texttt{YRBS} and \texttt{NSDUH} datasets.}
\vspace{-10pt}
\label{tab:performance}
\end{table*}
\paragraph{Bi-Modal Optimization.} 
To align modalities, we project the graph-based user embedding into the LLM embedding space via a learnable projection head $\mathbf{z}_u = P_{\text{proj}}(\mathbf{h}_u^{\text{agg}})$.
We prepend this projected graph token to the textual embeddings and feed the sequence into a frozen LLM to generate an output sequence $Y=(y_1,\ldots,y_r)$:
\begin{equation}
p_{\theta,\phi}(Y |  \mathcal{Z},\mathbf{z}_u)=\prod_{i=1}^{r}p_{\theta,\phi}\!\left(y_i | y_{<i},\,[\mathbf{z}_u;\mathcal{Z}]\right),
\end{equation}
where $\theta$ denotes frozen LLM parameters and $\phi$ includes trainable parameters of the graph encoder and $P_{\text{proj}}$. The concatenation $[\mathbf{z}_u;\mathcal{Z}]$ forms the full prompt, with $\mathbf{z}_u$ acting as a soft graph-conditioned prefix. In parallel, we feed $\mathbf{h}_u^{\text{agg}}$ to a lightweight classifier for illicit drug use prediction.

Let $y\in\{0,1\}$ be the ground-truth label and $\hat{y}$ the classifier output. We supervise the LLM with a verbalized label token (e.g., \texttt{Yes}/\texttt{No}) at the beginning of $Y$. The total loss combines a generation loss and a classification loss:

\begin{align}
    \nonumber \mathcal{L}_{\text{bi}} =\mathcal{L}_{\text{gen}} & +\mathcal{L}_{\text{cls}}, \quad \mathcal{L}_{\text{cls}}=\mathrm{CE}\big(\hat{y},y\big), \\
    \mathcal{L}_{\text{gen}} & =-\log p_{\theta,\phi}\!\left(y\mid \mathbf{z}_u,\mathcal{Z}\right). 
\end{align}
Gradients from $\mathcal{L}_{\text{bi}}$ update $\phi$ only. This dual-objective setup provides optimization signals from both textual space and graph space, enabling GNN-based structural reasoning and LLM-based textual interpretation.

\paragraph{Detection and Explanation.} 
At inference time, we use the graph classifier for illicit drug detection, and additionally prompt the LLM to justify the decision in natural language. This output serves as a post-hoc explanation, enabling qualitative analysis and uncovering interpretable patterns associated with illicit drug use.

\section{Experiments}

\begin{table*}[ht]
\centering
\small
\resizebox{\linewidth}{!}{
\begin{tabular}{l ccccc ccccc}
\toprule
\multirow{2}{*}{Model Variant} & \multicolumn{5}{c}{\texttt{YRBS}} & \multicolumn{5}{c}{\texttt{NSDUH}} \\
\cmidrule(lr){2-6} \cmidrule(lr){7-11}
 & Accuracy & Precision & Recall & F1-Score & AUC & Accuracy & Precision & Recall & F1-Score & AUC \\
\midrule
\textbf{LAMI} & \textbf{82.21} & \textbf{80.39} & \textbf{80.78} & \textbf{80.57} & \textbf{88.45} & \textbf{81.47} & \textbf{78.58} & \textbf{79.29} & \underline{78.86} & \textbf{88.23} \\ \midrule
\quad w/o Relation Matrix          & \underline{81.43} & 78.26 & 79.82 & 78.63 & 86.79 & \underline{80.44} & 76.33 & \underline{79.10} & 77.21 & 87.56 \\
\quad w/o Latent Relation Learning & 80.53 & 79.34 & 78.63 & 79.13 & \underline{87.29} & 79.73 & 77.12 & 79.05 & 77.80 & \underline{87.85} \\
\quad w/o \textsc{LLM}             & 80.06 & \underline{79.53} & \underline{80.44} & \underline{80.13} & 87.03 & 79.80 & \underline{78.22} & 79.04 & \textbf{78.91} & 87.74 \\
\bottomrule
\end{tabular}
}
\caption{Ablation study of \textbf{LAMI}. Each variant removes one key component from the full model.}
\vspace{-10pt}
\label{tab:ablation_lami}
\end{table*}

\subsection{Experiment Setup}

\paragraph{Baselines.}
We evaluate LAMI by comparing it against a diverse set of baseline models, including traditional machine learning methods (LightGBM~\cite{ke2017lightgbm} and XGBoost~\cite{chen2016xgboost}), neural network models for tabular data (MLP and TabNet~\cite{arik2021tabnet}), relational graph neural networks (RGCN~\cite{schlichtkrull2018modeling} and RGAT~\cite{busbridge2019relational}), and large language models (Qwen3-8B~\cite{yang2025qwen3}) evaluated in both zero-shot and LoRA~\cite{hu2022lora} settings. 

Notably, traditional machine learning algorithms such as LightGBM and XGBoost remain among the most competitive and robust baselines for learning from tabular data~\cite{shwartz2022tabular, borisov2022deep}, providing a strong benchmark for comparison.

\paragraph{Evaluation Settings.}
We evaluate our framework on both the \texttt{YRBS} and \texttt{NSDUH} datasets, using a standard data split of 70\% for training, 15\% for validation, and 15\% for testing. To comprehensively assess model performance across multiple dimensions, we report five widely used evaluation metrics: accuracy, precision, recall, macro-averaged F1-score, and ROC-AUC. All results are averaged over 10 runs with different random seeds to ensure robustness. For LAMI, we adopt Qwen3-0.6B as the LLM backbone to balance performance with computational feasibility, ensuring that the framework can be trained efficiently under available hardware resources.
The model is optimized using the Adam optimizer with a learning rate of  \(5 \times 10^{-5}\) and a weight decay of \(5 \times 10^{-4}\). Other key hyperparameters include: hidden dimension = 128, \(k_{\text{sim}} = 5\), \(k_{\text{att}} = 20\). 
Additional implementation details are provided in Appendix~\ref{sec:exp setting}. Prompts used are provided in Appendix~\ref{sec:prompt}

\subsection{Main Results}

Table~\ref{tab:performance} presents the performance of LAMI compared to several strong baselines across the \texttt{YRBS} and \texttt{NSDUH} datasets. On both datasets, LAMI consistently outperforms baselines across most evaluation metrics. Experiments show that RGCN and RGAT achieve performance comparable to, or in some metrics even surpassing, that of traditional machine learning methods, whereas MLP and TabNet underperform across the board. This highlights the effectiveness of our graph-based modeling approach in incorporating structural information into survey data analysis. In contrast, the relatively poor performance of the LLM-only model suggests that textual analysis alone is insufficient to uncover the underlying patterns embedded in the data. The strong results achieved by LAMI demonstrate that our framework effectively learns latent structures among nodes in the graph and leverages both structural and textual modalities to identify key substructures and make accurate predictions.

\subsection{Ablation Study}

\paragraph{The Impact of Model Components.}
Table~\ref{tab:ablation_lami} presents the ablation results of LAMI on both \texttt{YRBS} and \texttt{NSDUH}. Each variant removes one key component from the full model to assess its contribution. Specifically, \textit{LAMI (w/o Relation Matrix)} excludes the contextual relation matrices $W_r$ in the RGSL layer. \textit{LAMI (w/o Latent Relation Learning)} disables the latent relation learning entirely, thereby using only the original survey structure. \textit{LAMI (w/o LLM)} removes the LLM-based interpretability module, leaving only the graph-based classifier for prediction. The results show that each component contributes to overall performance improvement, underscoring the effectiveness of combining latent relation learning, contextualized structure learning, and language-based reasoning.

\begin{figure}[!t]
    \centering
    \includegraphics[width=\linewidth]{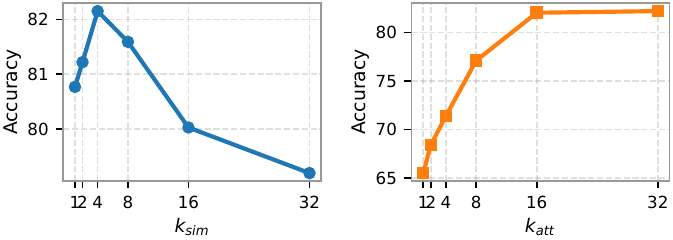}
    \vspace{-20pt}
    \caption{Model performance using different hyperparameter ($k_{sim}$ and $k_{att}$) settings.}
    \label{fig:hyper}
\end{figure}

\paragraph{The Impact of Inferred Relations.}
As shown in Figure~\ref{fig:hyper}, we how two key hyperparameters influence the performance of LAMI on \texttt{YRBS}: $k_{\text{sim}}$ and $k_{\text{att}}$. $k_{\text{sim}}$ controls how many \emph{new} latent relations the RGSL layer adds for each node during latent relation learning; $k_{\text{att}}$ controls how many questions are retained both for the graph-side classifier and for textualization to the LLM. For $k_{\text{sim}}$, accuracy initially increases as $k_{\text{sim}}$ grows, indicating that learned latent relations help the model capture useful dependencies that are not present in the original structure. However, performance degrades when too many new relations are added to the graph. This is because the inherent relations among question nodes are limited, and the noise induced by over-connecting the graph hinders the model from identifying key substructures. Increasing $k_{\text{att}}$ yields clear gains at smaller values since more informative substructures are provided to both the classifier and the LLM. Beyond approximately $k_{\text{att}}=20$, the improvement plateaus: the model has already captured the most predictive questions, and additional nodes contribute little to illicit-drug prediction while adding redundant information and longer prompts, leading to diminishing returns.

\begin{figure}[!t]
    \centering
    \includegraphics[width=\linewidth]{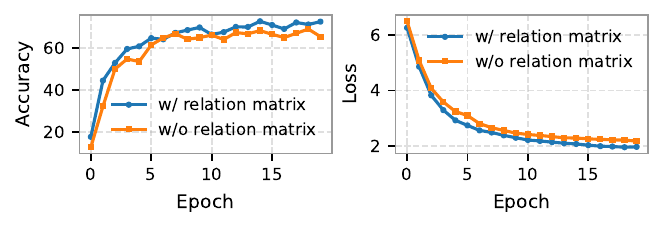}
    \vspace{-20pt}
    \caption{Training loss and model performance on \texttt{YRBS} with and without relation matrix in the latent relation learning stage.}
    \vspace{-10pt}
    \label{fig:ablation}
\end{figure}

\paragraph{The Impact of Relation Matrix.}
To further examine the effect of the relation matrix, Figure~\ref{fig:ablation} compares the training dynamics of the latent relation learning stage with and without relation matrices. Incorporating relation matrices leads to faster convergence and higher edge-type prediction accuracy, confirming that contextualized relation modeling provides more stable and informative structural learning.

\subsection{Interpretability Analysis}
\begin{figure*}
    \centering
    \includegraphics[width=\textwidth]{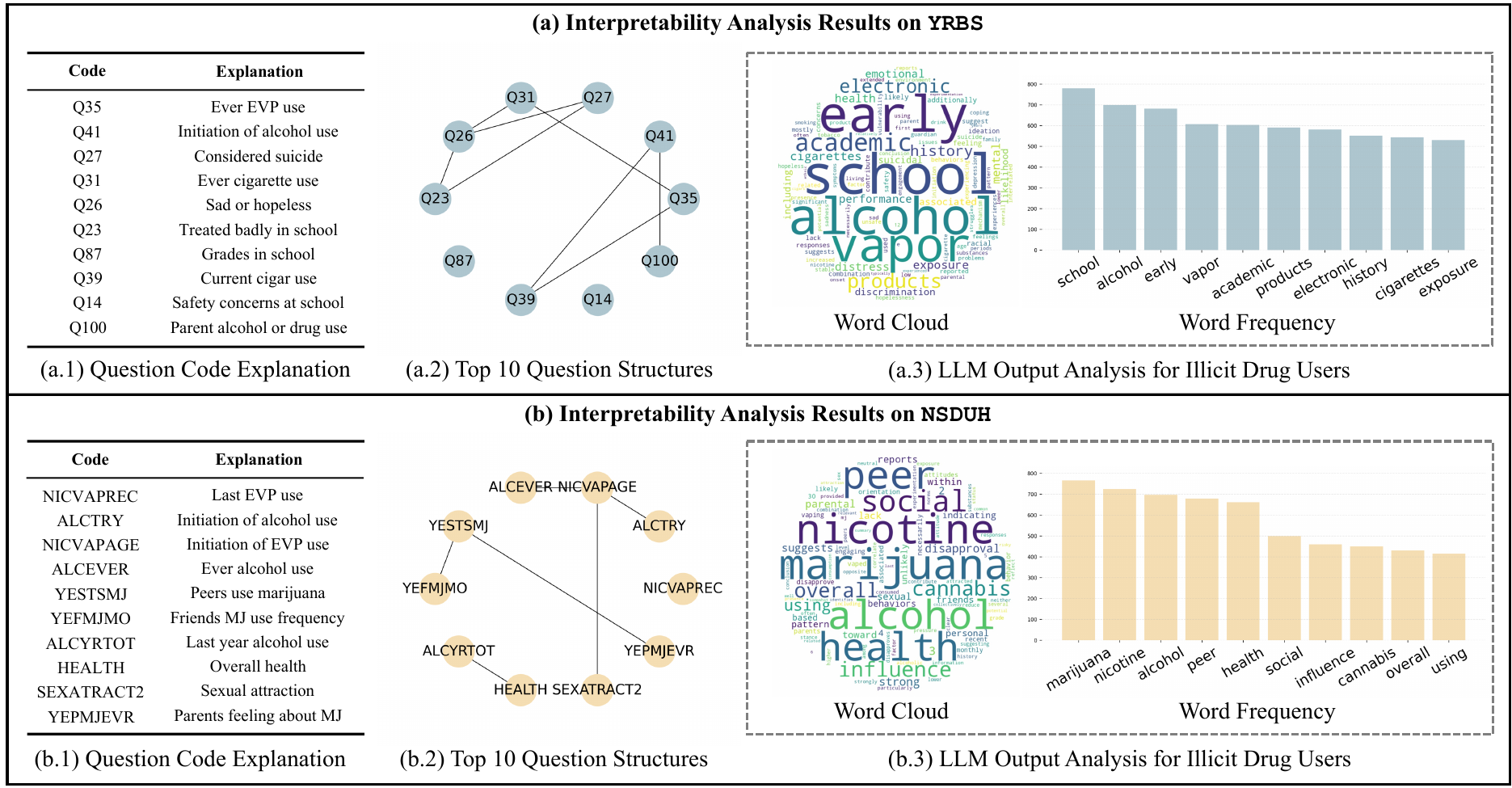}
    \vspace{-10pt}
    \caption{Interpretability analysis of LAMI on \texttt{YRBS} and \texttt{NSDUH} datasets, including (1) the top 10 contributing question nodes, (2) the learned structures among these nodes, and (3) the textual analysis based on the LLM's responses during inference.}
    \vspace{-10pt}
    \label{fig:interp}
\end{figure*}

To demonstrate the interpretability of LAMI, we visualize the top 10 questions based on their attention scores toward the user node, as illustrated in Figure~\ref{fig:interp}, and analyze their structural and semantic interrelations. For more detailed case studies on LLM outputs, please refer to Appendix~\ref{sec:case study}.

Despite survey design differences between the YRBS and NSDUH, several common themes emerge. Notably, early use of alcohol, nicotine, and electronic vapor products (EVPs) frequently appears among the most influential predictors. This observation is consistent with prior public health literature as well as the "gateway drug effect", showing that early initiation of legal substances is a significant risk factor for progression to illicit drug use~\cite{king2007prospective, trenz2012early}. Another recurrent theme is the impact of family environment. Parental substance use and negative parenting attitudes are strongly associated with increased likelihood of youth drug use. Previous studies have demonstrated that parental modeling, inconsistent discipline, and lack of emotional support are key risk factors for adolescent substance use~\cite{matejevic2014functionality, rusby2018influence}. 

Beyond individual items, LAMI captures meaningful inter-question dependencies. For instance, \texttt{Q35}–\texttt{Q39} and \texttt{Q41}–\texttt{Q39} edges reveal connections among different modes of tobacco and alcohol use, underscoring polysubstance patterns~\cite{crummy2020one}. More critically, LAMI discovers psychosocial pathways: \texttt{Q23} (Treated badly at school) links to \texttt{Q27} (Considered suicide), and both also connect to \texttt{Q26} (Felt sad or hopeless), forming a cluster that reveals the mental health consequences of school victimization. These psychological states are known precursors to substance use as maladaptive coping strategies~\cite{luk2010bullying}. \texttt{Q41} (Initiation of alcohol use) connects to \texttt{Q100} (Parental alcohol or drug use), reflecting the familial modeling of substance behavior~\cite{rusby2018influence}. 

In NSDUH, we observe analogous structures: \texttt{YESTSMJ} (Peer marijuana use) connects to \texttt{YEPMJEVR} (Parental view on marijuana) and \texttt{YEFMJMO} (Friend’s marijuana frequency), illustrating the compound social learning influence from both peers and family on youth~\cite{marziali2022perceptions}; \texttt{NICVAPAGE} (Initiation of vaping) is linked to \texttt{SEXATRACT2} (Sexual attraction), possibly reflecting minority stress pathways in LGBTQ+ youth, who exhibit higher substance use due to stigma, isolation, or mental distress~\cite{mann2021minority}.

In sum, LAMI not only improves predictive performance but also reveals complex interactions between early substance use, family dynamics, school environment, mental health, and peer influence. These insights can inform public health interventions and future research.

\section{Conclusion}
We introduce LAMI, a novel framework for illicit drug use prediction that models national survey datasets as relational graphs and learns implicit cross-question dependencies to enhance both prediction and interpretability. By combining Graph Neural Networks with Large Language Models, LAMI captures complex, latent structures while producing human-interpretable explanations. Empirical results on the \texttt{YRBS} and \texttt{NSDUH} datasets validate the superiority of our approach over strong baselines. Beyond quantitative performance, LAMI provides qualitative insights into the behavioral and psychosocial factors associated with illicit drug use among teenagers and young adults.

\section{Limitations}
While LAMI advances illicit drug use detection and interpretability on national surveys, several limitations remain. 
(i) \textbf{Dependence on survey topic design.} Our graph construction leverages survey-defined topic groupings; although these taxonomies are expert-informed, they may bias which cross-item relations are discoverable. Future work will explore data-driven or hybrid topic induction to yield more robust structures. 
(ii) \textbf{LLM capacity constraints.} Due to hardware limits, we employ a relatively small LLM backbone and keep its parameters frozen. Larger models may further improve both accuracy and explanation quality, but at the cost of compute and potential overfitting. 
(iii) \textbf{Temporal scope.} Experiments focus on the 2023 releases of YRBS and NSDUH data. Validating across multiple years and policy environments would better assess robustness to cohort and instrument shifts. 
(iv) \textbf{Faithfulness of explanations.} Our textual rationales are generated by an LLM conditioned on graph embeddings; although we supervise the answer token, the free-form explanations may not always be fully faithful to the decision process. Rigorous faithfulness metrics and expert evaluations are important next steps. 
(v) \textbf{Survey noise and missingness.} Self-reported outcomes can suffer from recall and social-desirability bias; skip logic and nonresponse introduce missingness patterns. Incorporating uncertainty modeling and missing-not-at-random strategies could increase reliability. 

\section{Ethical Considerations}

\paragraph{Data provenance and privacy.}
All analyses use the public-use microdata releases of NSDUH and YRBS, which are de-identified by the data providers and contain no direct identifiers. We follow the usage terms specified by SAMHSA and CDC, make no attempt at re-identification.

\paragraph{Responsible interpretation of LLM outputs.}
The natural-language rationales produced by our LLM are research aids rather than authoritative guidance. They may contain errors or omissions and should not be used for individual-level clinical, legal, or policy decisions without expert review. Our system is intended to support hypothesis generation and population-level analysis under human oversight.

\bibliography{main}
\normalsize

\clearpage
\appendix

\section{Data Preparation Details}
\label{sec:data prep}
In this section, we describe the construction of relational graphs and the labeling strategy for the YRBS and NSDUH survey datasets. 

\subsection{YRBS}
The official YRBS user guide\footnote{\url{https://www.cdc.gov/yrbs/data/index.html}} provides detailed information on survey design and codebooks. To define the binary label of illicit drug use, we use variables including \texttt{QNILLICT} and \texttt{QN46--55}. Respondents with a value of 1 in any of these variables are labeled as positive cases. For graph construction, we include only the raw survey questions (prefixed with \texttt{Q}), excluding their dichotomous counterparts (prefixed with \texttt{QN}) to avoid redundancy. The user node feature vector is derived from \texttt{Q1--7}, where \texttt{Q1--Q5} are one-hot encoded, and \texttt{Q6} (height) and \texttt{Q7} (weight) are normalized. To prevent information leakage, we exclude questions \texttt{Q46--55} and \texttt{Q92--93}, which directly indicate illicit drug use. Topic assignments are informed by the YRBS Explorer\footnote{\url{https://yrbs-explorer.services.cdc.gov/}} and the official handbook; we also employed an LLM-assisted approach to assign topics consistently.

\subsection{NSDUH}
The NSDUH user guide and codebook\footnote{\url{https://www.samhsa.gov/data/data-we-collect/nsduh-national-survey-drug-use-and-health/datafiles}} provide detailed documentation on survey methodology and variable design. The binary label of illicit drug use is defined using the variable \texttt{ILLYR}, with respondents coded as 1 classified as positive. NSDUH variables fall into three categories: (1) \emph{edited variables}, created after consistency checks; (2) \emph{imputed variables}, with missing values statistically imputed; and (3) \emph{recoded variables}, derived from edited or imputed sources. For graph construction, we use only edited variables to ensure consistency and minimize redundancy. User node features are built from \texttt{geographic} and \texttt{demographic} questions, one-hot encoded, with normalized height (\texttt{HTINCHE2}) and weight (\texttt{WTPOUND2}). To prevent information leakage, we exclude variables that directly indicate illicit drug use (e.g., \texttt{HPDRGTALK}, “Health professional discussed my drug use with me”). Topic definitions follow the codebook’s table of contents.

\section{Experimental Settings}
\label{sec:exp setting}
In addition to the hyperparameters described in the main paper, we set the batch size to 16 and train for 20 epochs during the Graph Latent Relation Learning stage. For the LLM-Enhanced Detection and Reasoning stage, we use a batch size of 4 and train for 10 epochs. For GNN-based baselines, we employ 3 layers and adopt the same hyperparameters as those used in LAMI wherever applicable.

\section{Prompt Design}

\label{sec:prompt}
\definecolor{myred}{RGB}{220,50,47}
\definecolor{mybrown}{RGB}{110,50,0}
\begin{figure}[h]
\centering
\begin{tcolorbox}[colback=myred!5!white,                
                  colframe=mybrown!80!white,
                  title= Prompt Templates]
{
\textbf{System Instruction:} \\
Here are some question-answer pairs provided by a user aged between 15 and 25 years old. Based on these facts, infer whether this user uses illicit drugs.

\medskip
\textbf{User Input Format:}
Here are the question-answer pairs:
\begin{verbatim}
Question: <question_1>  
Answer: <answer_1>
Question: <question_2>  
Answer: <answer_2>
...
\end{verbatim}
\textbf{Cues about learned relations:}
Think about the possible relations between \texttt{<question\_3>} and \texttt{<question\_4>} given the user's answers to them.
\begin{verbatim}
...
\end{verbatim}
\textbf{Following Prompt A:}
Based on these facts, can you infer whether this user uses illicit drugs?
Please answer with only "Yes" or "No".

\medskip
\textbf{Following Prompt B:}
Based on these facts, can you infer whether this user uses illicit drugs?
Please answer "Yes" or "No" and explain why.
}
\end{tcolorbox}
\vspace{-10pt}
\caption{The shared prompt templates we used for experiments.}
\vspace{-10pt}
\end{figure}
\label{fig:prompt}

We use the prompt templates shown in Figure~\ref{fig:prompt}. When finetuning and evaluating Qwen3-8B with LoRA and evaluating Qwen3-8B under zero-shot setting, we exclude \textit{Cues about learned relations} and adopt \textit{Following Prompt A}. When training LAMI, we adopt \textit{Following Prompt A} to make it easier for computing $\mathcal{L}_{\text{gen}}$. While in inference time, we adopt \textit{Following Prompt B} to obtain justified textual output from the LLM module to facilitate further analysis.

\section{Case Study}
\label{sec:case study}

To qualitatively assess how LAMI integrates behavioral and psychosocial information for prediction and explanation, we present two representative cases. Case~A, sampled from the NSDUH dataset, represents a high-risk individual, while Case~B, sampled from the YRBS dataset, reflects a low-risk profile. Each case includes the user’s survey-derived question–answer pairs, the latent relation cues automatically provided by the model, and the final text-based inference generated by the LLM module.

\paragraph{Case A: High-Risk Profile.}
Case~A describes a respondent exhibiting multiple behavioral and psychosocial risk factors. The user reports recent cigarette and e-cigarette use, low parental supervision, negative attitudes toward school, moderate academic performance, and several depressive symptoms such as fatigue, low energy, and feelings of worthlessness. The individual also lives in poverty and reports having carried a handgun once or twice. These variables collectively depict an environment characterized by psychosocial stress, weak family regulation, and exposure to risk-taking peers—conditions widely recognized in the literature as predictors of substance use.

LAMI’s reasoning mirrors these patterns. The learned relation cues emphasize connections between parental monitoring and peer drinking behavior, between poverty and depressive affect, and between low school engagement and vaping. These relations correspond to established mechanisms such as social learning and stress–coping pathways~\cite{trucco2020review, king2008adolescent}. The model’s textual output correctly infers illicit drug use, highlighting low parental control, peer influence, and emotional distress as interacting factors contributing to elevated risk. This demonstrates that LAMI can synthesize structural and contextual cues into an interpretable, human-like explanation grounded in behavioral science theory.

\paragraph{Case B: Low-Risk Profile.}
Case~B represents a user with minimal behavioral risk indicators. The individual reports no cigarette, vaping, or alcohol use, no experiences of bullying or violence, and no depressive symptoms. Although the user indicates limited parental awareness, low physical activity, and modest academic performance, these factors alone are insufficient to suggest substance use. Overall, the profile reflects a relatively stable psychosocial environment with low direct exposure to drugs or associated behaviors.

The latent relation cues in this case focus on benign correlations—such as the link between limited parental supervision and physical inactivity, or between sleep duration and academic outcomes—none of which directly imply substance use risk. Consistent with these signals, LAMI predicts a negative outcome (\texttt{No}) and produces a concise rationale noting the absence of substance-related behaviors and major psychosocial distress. The model’s explanation demonstrates appropriate calibration: it recognizes secondary vulnerabilities but does not overgeneralize them into a positive prediction.

\paragraph{Summary.}
Together, these cases illustrate that LAMI not only achieves accurate classification but also generates contextually grounded explanations aligned with established psychosocial mechanisms. In high-risk scenarios, the model integrates cross-domain cues—such as peer exposure, parental disengagement, and emotional dysregulation—to justify its prediction. In low-risk cases, it maintains conservative reasoning, acknowledging minor behavioral concerns without inferring unsupported drug use. These results underscore LAMI’s interpretability and its potential to aid behavioral scientists in understanding the multifactorial nature of youth substance use.

\begin{figure*}[h]
\begin{tcolorbox}[colback=myred!5!white,                
                  colframe=mybrown!80!white,
                  title= Case A: input (part I)]
\textbf{System Instruction:} \\
Here are some question-answer pairs provided by a user aged between 15 and 25 years old. Based on these facts, infer whether this user uses illicit drugs.

\medskip
\textbf{User Input:}
\begin{verbatim}
Question: EVER SMOKED A CIGARETTE 
Answer: 1 - Yes
Question: TIME SINCE LAST SMOKED CIGARETTES 
Answer: 1 - Within the past 30 days
Question: AVG # CIGS SMOKED PER DAY/ON DAY SMOKED 
Answer: 3 - 2 to 5 cigarettes per day
Question: EVER VAPED NICOTINE WITH E-CIGARETTE OR OTHER VAPING DEVICE 
Answer: 1 - Yes
Question: TIME SINCE LAST VAPED NICOTINE WITH E-CIG OR VAPING DEVICE 
Answer: 1 - Within the past 30 days
Question: HOW MANY SDNTS YOU KNOW IN GRADE SMOKE CIGARETTES 
Answer: 3 - Most of them
Question: HOW MANY SDNTS YOU KNOW GET DRUNK WEEKLY 
Answer: 3 - Most of them
Question: HOW YOU FELT OVERALL ABT GOING TO SCHL PAST 12 MOS 
Answer: 4 - You hated going to school
Question: HOW OFTEN FELT SCHL WORK MEANINGFUL PST 12 MOS 
Answer: 4 - Never
Question: GRADES FOR LAST SEMESTER/GRADING PERIOD COMPLETED 
Answer: 3 - A C+ , C or C-minus average
Question: PRNTS CHECK IF YOU'VE DONE HOMEWORK PST 12 MOS 
Answer: 4 - Never
Question: PRNTS LIMITED YOUR TIME OUT W/FRNDS PST 12 MOS 
Answer: 4 - Never
Question: PRNTS LET YOU KNOW DONE A GOOD JOB PST 12 MOS 
Answer: 4 - Never
Question: ARGUED/HAD A FIGHT WITH AT LEAST ONE OF YOUR PRNTS 
Answer: 4 - 6 to 9 times
Question: SCORE OF SYMPTOM INDICATORS 1 THRU 9 
Answer: 1 - Has 5 or more symptoms
Question: FELT WORTHLESS NEARLY EVERY DAY 
Answer: 1 - Has symptom
Question: FELT TIRED/LOW ENERGY NEARLY EVERY DAY 
Answer: 1 - Has symptom
Question: INABILITY TO CONCENTRATE OR MAKE DECISIONS 
Answer: 1 - Has symptom
Question: RC-POVERTY LEVEL-NEW INC (% OF US CENSUS POVERTY THRESHOLD) 
Answer: 1 - Living in Poverty
Question: CARRIED A HANDGUN 
Answer: 2 - 1 or 2 times
\end{verbatim}
\end{tcolorbox}
\caption{Case A user input showing a high-risk behavioral and psychosocial profile.}
\end{figure*}
\label{fig:case a 1}

\begin{figure*}[h]
\begin{tcolorbox}[colback=myred!5!white,                
                  colframe=mybrown!80!white,
                  title= Case A: Input (Part II)]
\textbf{Cues about learned relations:}
Think about the possible relations between \texttt{PRNTS LIMITED YOUR TIME OUT W/FRNDS PST 12 MOS} and \texttt{HOW MANY SDNTS YOU KNOW GET DRUNK WEEKLY} given the user's answers to them.\\
Think about the possible relations between \texttt{RC-POVERTY LEVEL-NEW INC (\% OF US CENSUS POVERTY THRESHOLD)} and \texttt{FELT WORTHLESS NEARLY EVERY DAY} given the user's answers to them.\\
Think about the possible relations between \texttt{EVER VAPED NICOTINE WITH E-CIGARETTE OR OTHER VAPING DEVICE} and \texttt{HOW YOU FELT OVERALL ABT GOING TO SCHL PAST 12 MOS} given the user's answers to them.\\
Think about the possible relations between \texttt{CARRIED A HANDGUN} and \texttt{PRNTS LET YOU KNOW DONE A GOOD JOB PST 12 MOS} given the user's answers to them.\\
Think about the possible relations between \texttt{HOW MANY SDNTS YOU KNOW IN GRADE SMOKE CIGARETTES} and \texttt{INABILITY TO CONCENTRATE OR MAKE DECISIONS} given the user's answers to them.

\textbf{Following Prompt:}
Based on these facts, can you infer whether this user uses illicit drugs?
Please answer "Yes" or "No" and explain why.
\end{tcolorbox}
\caption{Learned latent relations used as reasoning cues in Case A’s LLM prompt.}
\end{figure*}
\label{fig:case a 2}

\begin{figure*}[h]
\begin{tcolorbox}[colback=myred!5!white,                
                  colframe=mybrown!80!white,
                  title= Case A: Model Output]
\textbf{Model output:} Yes. The user reports recent cigarette and vaping use, low parental monitoring, poor school engagement, and several depressive symptoms, which are common indicators linked to higher drug-use risk. The user also has peers who drink frequently, which may reinforce such behaviors through social influence. Considering these patterns together, especially the link between low parental control and strong peer exposure, I think the answer is yes.
\end{tcolorbox}
\caption{Model inference and explanation for the high-risk Case A user.}
\end{figure*}
\label{fig:case a output}

\begin{figure*}[h]
\begin{tcolorbox}[colback=myred!5!white,                
                  colframe=mybrown!80!white,
                  title= Case B: Input (Part I)]
\textbf{System Instruction:} \\
Here are some question-answer pairs provided by a user aged between 15 and 25 years old. Based on these facts, infer whether this user uses illicit drugs.

\medskip
\textbf{User Input:}
\texttt{
Question: Have you ever smoked a cigarette, even one or two puffs? 
Answer: No
Question: During the past 30 days, on how many days did you smoke cigarettes? 
Answer: 0 days
Question: Have you ever used an electronic vapor product? 
Answer: No
Question: During the past 30 days, on how many days did you use an electronic vapor product? 
Answer: 0 days
Question: How old were you when you had your first drink of alcohol other than a few sips? 
Answer: Never drank alcohol
Question: During the past 30 days, on how many days did you have at least one drink of alcohol? 
Answer: 0 days
Question: Did you drink alcohol or use drugs before you had sexual intercourse the last time? 
Answer: No
Question: During the past 12 months, did you ever feel so sad or hopeless almost every day for two weeks or more in a row that you stopped doing some usual activities? 
Answer: No
Question: During the past 12 months, did you ever seriously consider attempting suicide? 
Answer: No
Question: During the past 12 months, have you ever been bullied on school property? 
Answer: No
Question: During the past 12 months, have you ever been electronically bullied? (Count being bullied through texting, Instagram, Facebook, or other social media.) 
Answer: No
Question: During the past 12 months, how many times were you in a physical fight? 
Answer: 0 times
Question: During the past 12 months, how many times did someone threaten or injure you with a weapon such as a gun, knife, or club on school property? 
Answer: 0 times
Question: How often do your parents or other adults in your family know where you are going or with whom you will be? 
Answer: Never
Question: Do you agree or disagree that you feel close to people at your school? 
Answer: Disagree
Question: On an average school night, how many hours of sleep do you get? 
Answer: 5 hours
Question: During the past 7 days, on how many days were you physically active for a total of at least 60 minutes per day? (Add up all the time you spent in any kind of physical activity that increased your heart rate and made you breathe hard some of the time.) 
Answer: 0 days
Question: During the past 12 months, how would you describe your grades in school? 
Answer: Mostly C's
Question: During your life, how often has there been an adult in your household who tried hard to make sure your basic needs were met, such as looking after your safety and making sure you had clean clothes and enough to eat? 
Answer: Never
Question: When was the last time you saw a dentist for a check-up, exam, teeth cleaning, or other dental work? 
Answer: Never
}
\end{tcolorbox}
\caption{Case B user input showing a low-risk behavioral and health profile.}
\end{figure*}
\label{fig:case b 1}

\begin{figure*}[h]
\begin{tcolorbox}[colback=myred!5!white,                
                  colframe=mybrown!80!white,
                  title= Case B: Input (Part II)]

\textbf{Cues about learned relations:}
Think about the possible relations between \texttt{How often do your parents or other adults in your family know where you are going or with whom you will be?} and \texttt{During the past 7 days, on how many days were you physically active for a total of at least 60 minutes per day?} given the user's answers to them.\\
Think about the possible relations between \texttt{On an average school night, how many hours of sleep do you get?} and \texttt{During the past 12 months, how would you describe your grades in school?} given the user's answers to them.\\
Think about the possible relations between \texttt{During the past 30 days, on how many days did you have at least one drink of alcohol?} and \texttt{Did you drink alcohol or use drugs before you had sexual intercourse the last time?} given the user's answers to them.\\
Think about the possible relations between \texttt{During your life, how often has there been an adult in your household who tried hard to make sure your basic needs were met, such as looking after your safety and making sure you had clean clothes and enough to eat?} and \texttt{When was the last time you saw a dentist for a check-up, exam, teeth cleaning, or other dental work?} given the user's answers to them.\\
Think about the possible relations between \texttt{During the past 12 months, have you ever been electronically bullied? (Count being bullied through texting, Instagram, Facebook, or other social media.)} and \texttt{During the past 12 months, did you ever feel so sad or hopeless almost every day for two weeks or more in a row that you stopped doing some usual activities?} given the user's answers to them.

\textbf{Following Prompt:}
Based on these facts, can you infer whether this user uses illicit drugs?
Please answer "Yes" or "No" and explain why.
\end{tcolorbox}
\caption{Learned latent relations used as reasoning cues in Case B’s LLM prompt.}
\end{figure*}
\label{fig:case b 2}

\begin{figure*}[h]
\begin{tcolorbox}[colback=myred!5!white,                
                  colframe=mybrown!80!white,
                  title= Case B: Model Output]
\textbf{Model output:} No. They report no cigarette, vaping, or alcohol use, and also no use before sex. No fights or bullying and no recent sadness. There are some risks like less sleep, low activity, weaker grades, and low parental monitoring, but overall there isn’t evidence of illicit drug use here.
\end{tcolorbox}
\caption{Model inference and explanation for the low-risk Case B user.}
\end{figure*}
\label{fig:case b output}

\end{document}